\documentclass{article}
\usepackage{PRIMEarxiv}

\usepackage{hyperref}
\usepackage{amsmath,amssymb,amsfonts}
\usepackage{algorithmic}
\usepackage{graphicx}
\usepackage{textcomp}
\usepackage{xcolor}
\usepackage{multirow}

\title{Knowledge Distillation of Domain-adapted LLMs for Question-Answering in Telecom\\
}

\author{Rishika Sen,  Sujoy Roychowdhury,  Sumit Soman,  H. G. Ranjani, Srikhetra Mohanty \\
Ericsson R\&D, Bangalore, India \\
\{rishika.sen, sujoy.roychowdhury, sumit.soman, ranjani.h.g, srikhetra.mohanty\}@ericsson.com}

\begin{document}
\maketitle

\begin{abstract}
Knowledge Distillation (KD) is one of the approaches to reduce the size of Large Language Models (LLMs). A LLM with smaller number of model parameters (student) is trained to mimic the performance of a LLM of a larger size (teacher model) on a specific task. For domain-specific tasks, it is not clear if teacher or student model, or both, must be considered for domain adaptation. In this work, we study this problem from perspective of telecom domain Question-Answering (QA) task. We systematically experiment with Supervised Fine-tuning (SFT) of teacher only, SFT of student only and SFT of both prior to KD. We design experiments to study the impact of vocabulary (same and different) and KD algorithms (vanilla KD and Dual Space KD, DSKD) on the distilled model. Multi-faceted evaluation of the distillation using 14 different metrics (N-gram, embedding and LLM-based metrics) is considered. Experimental results show that SFT of teacher improves performance of distilled model when both models have same vocabulary, irrespective of algorithm and metrics. Overall, SFT of both teacher and student results in better performance across all metrics, although the statistical significance of the same depends on the vocabulary of the teacher models.  

\end{abstract}

\section{Introduction}
Large Language Models (LLMs) are complex models that perform a wide range of tasks, while Small Language Models (SLMs) have fewer parameters and are more suited for specific, resource-constrained applications. It has been well established that domain adaptation improves performance of LLMs in technical domains, such as telecom \cite{soman2023observations,  bariah2023understanding, roychowdhury2024evaluation, thanos2024, zou2024telecomgpt}. The need for SLMs arises due to their efficiency and cost-effectiveness as against LLMs \cite{piovesan2024telecom, maatouk2024large, schick2020s}. Techniques to reduce the size of LLMs while retaining much of their performance is an area of active research. Popular techniques include quantization \cite{zhang2023revisiting}, 
pruning \cite{ma2023llm} and Knowledge Distillation (KD) \cite{gou2021knowledge}. 

In this work, we focus on the impact of domain adaptation of LLMs via KD approach. KD is a technique where a ``student" (smaller) model is trained to replicate the performance of a ``teacher" (larger) model \cite{gou2021knowledge, xu2024survey} for a particular task. KD was originally proposed to reduce model size while retaining performance \cite{Hinton2015DistillingTK}.

\begin{figure}[h]
\centering
\includegraphics[width=0.98\textwidth]{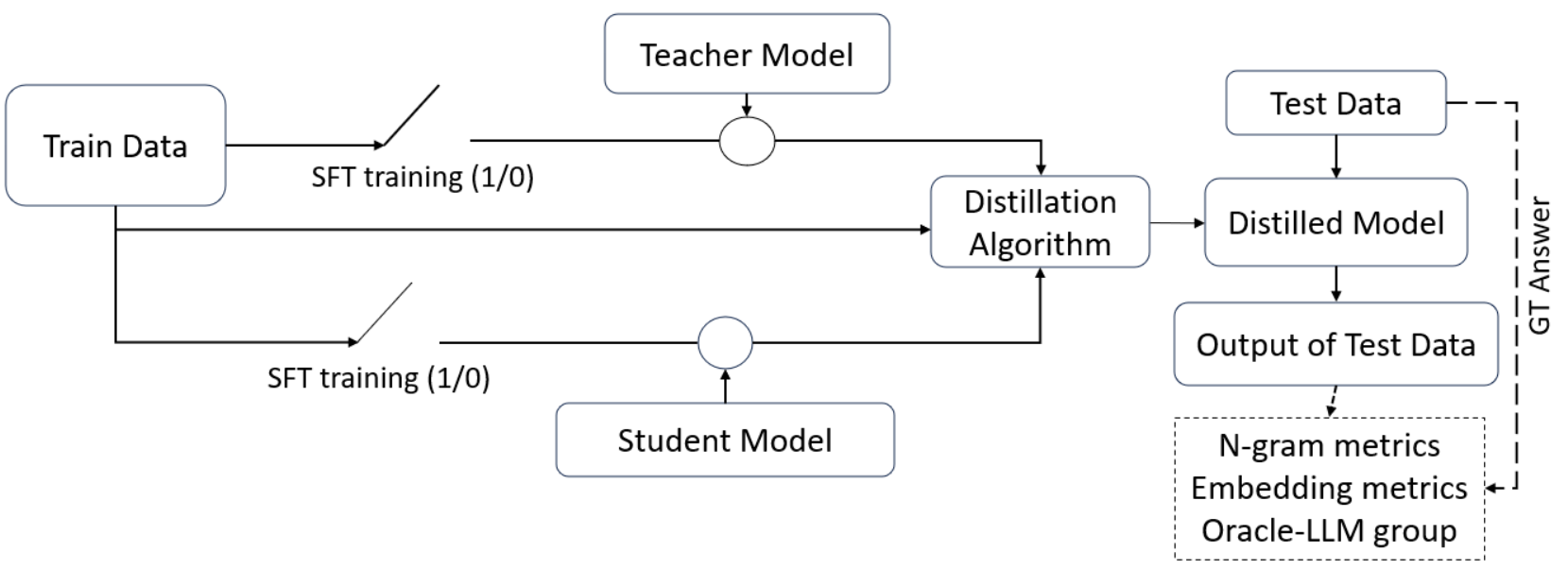}
\caption{A schematic representation of experiments consisting of the choice of SFT for teacher student, the choice of distillation algorithms, Vanilla or DSKD, and choice of evaluation metrics.}
\label{pic:switch}
\end{figure}

\subsection{Problem statement}
The smaller models obtained during KD is said to improve generalization, reduce overfitting, especially when trained on small datasets. These small models enable faster inference and lower deployment cost. 
In domain specific tasks, such as telecom, it is important to ensure the models considered are domain aware. This is typically achieved through pre-training and/or supervised fine-tuning (SFT). SFT is a model training technique where a pre-trained model is further trained on a labeled dataset via supervised learning \cite{vaswani2017attention}.  

To the best of our knowledge, there has been no work pertaining to impact of domain adaptation through SFT of either teacher or student models prior to distillation. In addition, there are no insights on how one must choose the teacher and student models \textit{viz.} must they be of same vocabulary or different. Lastly, quantifying performance of distilled generative models requires a holistic evaluation \cite{roychowdhury2024evaluation} as considering just N-gram metrics or embedding based similarity metrics can be severely limiting for LLMs. \\

Hence, in-lieu of these gaps, we formalize the research questions in this work as follows: 
\begin{itemize}
    \item \textbf{RQ1:} Does SFT of teacher and/or student models prior to KD improve distilled model performance? 
    \item \textbf{RQ2:} Does the choice of models for SFT and KD impact performance i.e., are there advantages in using models of different vocabulary over same vocabulary?
    \item \textbf{RQ3:} Does performance change for different metric groups - N-gram based, embedding based and Oracle-LLM based metrics?
\end{itemize}

\subsection{Overview of KD techniques}
\label{ssub:kd_overview}
Vocabulary of the models chosen for KD impact the performance of the distilled model. In this sub-section, we give a quick overview of vocabulary prior to overview of KD techniques in literature.  
\subsubsection{Vocabulary} LLM vocabulary refers to the set of tokens (comprising of words, sub-words, or characters) that is used to represent text \cite{kolesnikova2022knowledge}. Tokenization is the process of converting input text into tokens, and subsequently converted to embedding vectors. Tokenization and thus vocabulary plays an important role in model performance. It is evident that the vocabularies across LLM families differ based on the tokenization \cite{kolesnikova2022knowledge}. 

\subsubsection{KD Algorithms} Typical KD training process involves comparing the token representations of the teacher and student models to align the latter to that of former using KL divergence (KLD) loss \cite{aguilar2020knowledge}. We refer to this technique, in this paper, as vanilla KD. 

  When teacher and student model have different vocabulary, the models predict the tokens of next word in sequence from different vocabularies \cite{kolesnikova2022knowledge}. Thus, it can be expected that vocabulary plays an important role in distilled model performance. Performance improvement in the scenario where the vocabularies do not match has been addressed by extending the vanilla KD approach by projecting each model's embeddings to a unified space; this is referred to Dual Space KD or DSKD \cite{zhang2024dual}. For creating the unified space, the outputs from teacher model space are projected on to student model space and vice-versa. The transformation for projection is learnt during Cross-Model Attention (CMA) \cite{zhang2024dual} mechanism which also bridges the difference in vocabularies.

In this work, we consider two models (Llama family and Mistral family) and two KD approaches - vanilla KD and DSKD - to study the impact of vocabulary and algorithms on domain adaptation of LLMs. To the best of our knowledge, our work is the first to study the effect of domain adaptation through SFT of both the teacher and the student LLMs prior to distillation.
\vspace{-2mm}
\subsection{Overview of metrics}
\label{ssub:metrics_overview}
Evaluation of generated output from a LLM is an evolving research topic \cite{desmond2024evalullm, roychowdhury2024evaluation}. The current KD approaches typically report on few N-gram based metrics only \cite{zhang2024dual}. For a more rounded evaluation of LLM, we consider three metric groups - N-gram based metrics, embedding based metrics and Oracle-LLM based metrics. The specific metrics are listed below:
\begin{itemize}
    \item \textbf{N-gram based metrics}: BLEU, BLEU-CN , BLEU-DM , BLEU-DC \cite{shi2022evaluation}, ROUGE-L Precision and Recall \cite{lin-2004-rouge}. These scores are indicative of overlap of N-gram word sequences or longest common sequences. 
    \item \textbf{Embedding based metrics}: Cosine similarity (using all-Mini-L6-v2 embeddings \cite{reimers2020making}), BERTScore  \cite{zhang2019bertscore}. These scores are indicative of semantic similarity. 
    \item \textbf{Oracle-LLM based metrics}: RAG Assessment metrics (RAGAs) that uses Oracle-LLM to arrive at metrics such as faithfulness, factual correctness, answer similarity, answer correctness, answer relevance and context relevance,  \cite{es2024ragas}, \cite{roychowdhury2024evaluation}. 

\end{itemize}
Higher scores imply better model performance for all the metrics considered above.  
Capturing KD performance using set of metrics which cover word/token overlap, semantic similarity and generation perspective aids towards holistic  analysis. 

\subsection{Contributions}
\label{ssub:contributions}
From the experiments designed and through the results on TeleQuAD \cite{telequad2025}, the contributions of this work are:

\begin{itemize}
    \item This is the first work which addresses the effect of supervised fine tuning (SFT) of both the teacher and the student language models prior to distillation (with a focus on telecom domain QA task).
\item We demonstrate that SFT of teacher and student models improves performance, irrespective of vocabulary and algorithm choice. 
\item We demonstrate SFT of teacher has significant performance improvements (across metrics) when using same vocabulary models. 
\item In scenarios where SFT training has practical limitations, using different vocabulary with DSKD algorithm is found to be useful. 
    \item All group-wise metrics show similar performance trends. 
\end{itemize}
The rest of the paper is organized as follows. The experimental design and evaluation metrics are described in Section \ref{sec:Methodology}, followed by experimental setup and results in Section \ref{sec:ExperimentalResults}. We conclude and discuss future work in Section \ref{sec:conclusions}.

\section{Methodology}\label{sec:Methodology}
We describe the experimental setup, statistical tests and the details of dataset and models.
\begin{figure*}[h!]
\centering
\includegraphics[width=0.75\textwidth]{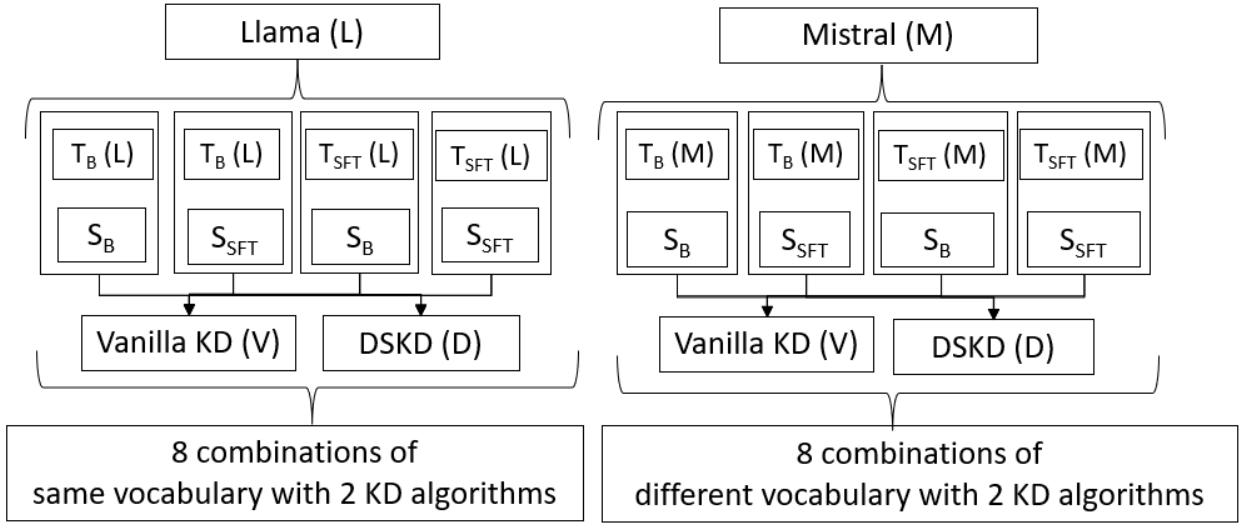}
\caption{Schematic representation of different choices based on which we conduct Hypothesis tests}
\label{pic:statTests}
\end{figure*}
\subsection{Experimental Setup}
\label{subsec:experiments}
We describe the experimental design considered to study the impact of vocabulary (same and different) and KD
algorithms (vanilla KD and Dual Space KD, DSKD) on the distilled model. We also analyze the impact of untrained/SFT teacher/student model on the final distilled model. Fig. \ref{pic:switch} shows the schematic representation of our experimental setup. Depending on choice of SFT training, teacher model and distillation algorithms, there are 4 parameters of interest here:
\begin{itemize}
    \item \textbf{Teacher} – Two variants of the teacher model, base model and SFT model, arising out of SFT training (depicted as a 1/0 switch) in Fig. \ref{pic:switch}. 
    \item \textbf{Student} – Similarly, student model is also considered with two variants - base model and SFT model (refer to Fig. \ref{pic:switch}). These two addresses RQ1. 
    \item \textbf{Vocabulary} – To study impact of vocabulary, we consider two cases where teacher and student (i) both having same vocabulary (ii) both having different vocabulary. We fix student model to be from the Llama family - TinyLlama. Hence, with respect to the student SLM TinyLlama, choosing the teacher model as (i) LLM Llama results in same vocabulary (ii) LLM Mistral results in different vocabulary. This addresses RQ2.
    \item \textbf{KD algorithm} – To analyze if insights on vocabulary is invariant to choice of KD algorithm, we consider two KD algorithms - Vanilla KD and DSKD. 
\end{itemize}
{\color{black} Fig. \ref{pic:statTests} shows a schematic representation of 16 combinations of experiments arising out of the design described above. The notations used to depict the various combinations are summarized in Table \ref{tab:notation}.  We report performance using 14 metrics (refer Section \ref{ssub:metrics_overview}) for each of the 16 combinations of distillation experiments.

\begin{table}[h]
\centering
\begin{tabular}
{|p{0.15\columnwidth}|p{0.75\columnwidth}|}
\hline
Notation & Description\\\hline\hline
    $T_B(L)$ & Base Llama as teacher model (Same vocabulary) \\
    $T_{SFT}(L)$& SFT Llama as teacher model (Same vocabulary)\\
    $T_B(M)$ & Base Mistral as teacher model (Different vocabulary)\\
    $T_{SFT}(M)$& SFT Mistral as teacher model (Different vocabulary) \\
    $S_B$ & Base TinyLlama as student model\\
    $S_{SFT}$& SFT TinyLlama as student model \\
    $V$ and $D$ & Vanilla algorithm and DSKD algorithm for KD process \\\hline
\end{tabular}\vspace{1mm}
\caption{A summary of notations used to formulate hypothesis tests and report results.}
\label{tab:notation}
\end{table}
\subsection{Hypothesis tests}
\label{ssub:hypothesis_tests}

In addition to reporting the performance metrics, we analyze the impact of the SFT of teacher, student, model vocabulary and the algorithm chosen for each metric. This results in 16 combinations of results for 14 metrics (RQ3); to ensure the results are statistically significant, we group the results to perform statistical hypothesis tests (Wilcoxon statistics signed rank test \cite{gehan1965generalized}). 

We henceforth use the notation where a tuple ($T_B$, $S_B$) indicates a teacher-student pair where the models are identified by the acronyms as in Table \ref{tab:notation}. Using a wildcard * in the suffixes indicate all possible options of the latter. When we use a third term in the tuple e.g. D or V we refer to the corresponding algorithm - not having this indicates that we test for both algorithms. We compare each of the 14 metrics via a statistical test and this is indicated by the $Perf()$ function.

\begin{itemize}
    \item \textbf{H-Train}: We consider the null hypotheses (NH), Eq. (\ref{eqn:rq3_1}), to analyze if $T_{SFT}$ or $S_{SFT}$ or both (followed by KD) impacts the performance of the distilled model (RQ1).
 \begin{gather}
    H_{train}^{T}: Perf(T_{B}, S_B) = Perf(T_{SFT}, S_B) \nonumber \\
    H_{train}^{S}: Perf(T_{B}, S_B) = Perf(T_{B}, S_{SFT}) \nonumber \\
    H_{train}^{T,S}: Perf(T_{B}, S_B) = Perf(T_{SFT}, S_{SFT}) \label{eqn:rq3_1}
\end{gather}
The alternate hypotheses to all of these correspond to $Perf(T_B, S_B) \ne Perf(T_*, S_*)$ respectively where $*$ corresponds to the SFT of teacher or student or both. 
For each of the three NH above, impact on vocabulary ($L$ and $M$) and algorithm ($V$ and $D$) choice are also considered. So, we have $12$ hypothesis tests for each of the 14 metrics. 
    \item \textbf{H-SFT}: Results show that $T_{SFT},S_{SFT}$ combination results in best performance across metrics (discussed later in Section \ref{sec:ExperimentalResults}). To analyze the impact of the $T_{SFT}$ only, $S_{SFT}$ only and $(T_{SFT}, S_{SFT})$ prior to the distillation process (RQ1), we formulate NH as Eq. (\ref{eqn:rq3_2}).
     \begin{gather}
    H_{SFT}^{T}: Perf(T_{SFT},S_{SFT}) = Perf(T_{SFT},S_{B}) \nonumber \\
    H_{SFT}^{S}: Perf(T_{SFT},S_{SFT}) = Perf(T_{B},S_{SFT}) \label{eqn:rq3_2}
\end{gather}
Again, the alternate hypotheses to all of these correspond to $Perf(T_{SFT}, S_{SFT}) \ne Perf(T_*, S_*)$ respectively. Here, $*$ refers to SFT of teacher model only or student model only. 
Each of the binary choice of vocabulary ($M$, $L$) and algorithm ($V$, $D$) is considered resulting in $8$ tests for each of the 14 metrics. 
\item \textbf{H-Algo}: Impact of KD algorithm (RQ2) post SFT through NH is shown in Eq. (\ref{eqn:rq3_4}).
\begin{gather}
    H_{Alg}^{T,S}: Perf(T_{SFT}, S_{SFT}, V) = Perf(T_{SFT}, S_{SFT}, D)\nonumber\\
    H_{Alg}^B: Perf(T_{B}, S_{B}, V) = Perf(T_{B}, S_{B}, D)\label{eqn:rq3_4}
\end{gather}

\begin{figure*}[h!]
{
\centering
\includegraphics[width=0.95\textwidth]{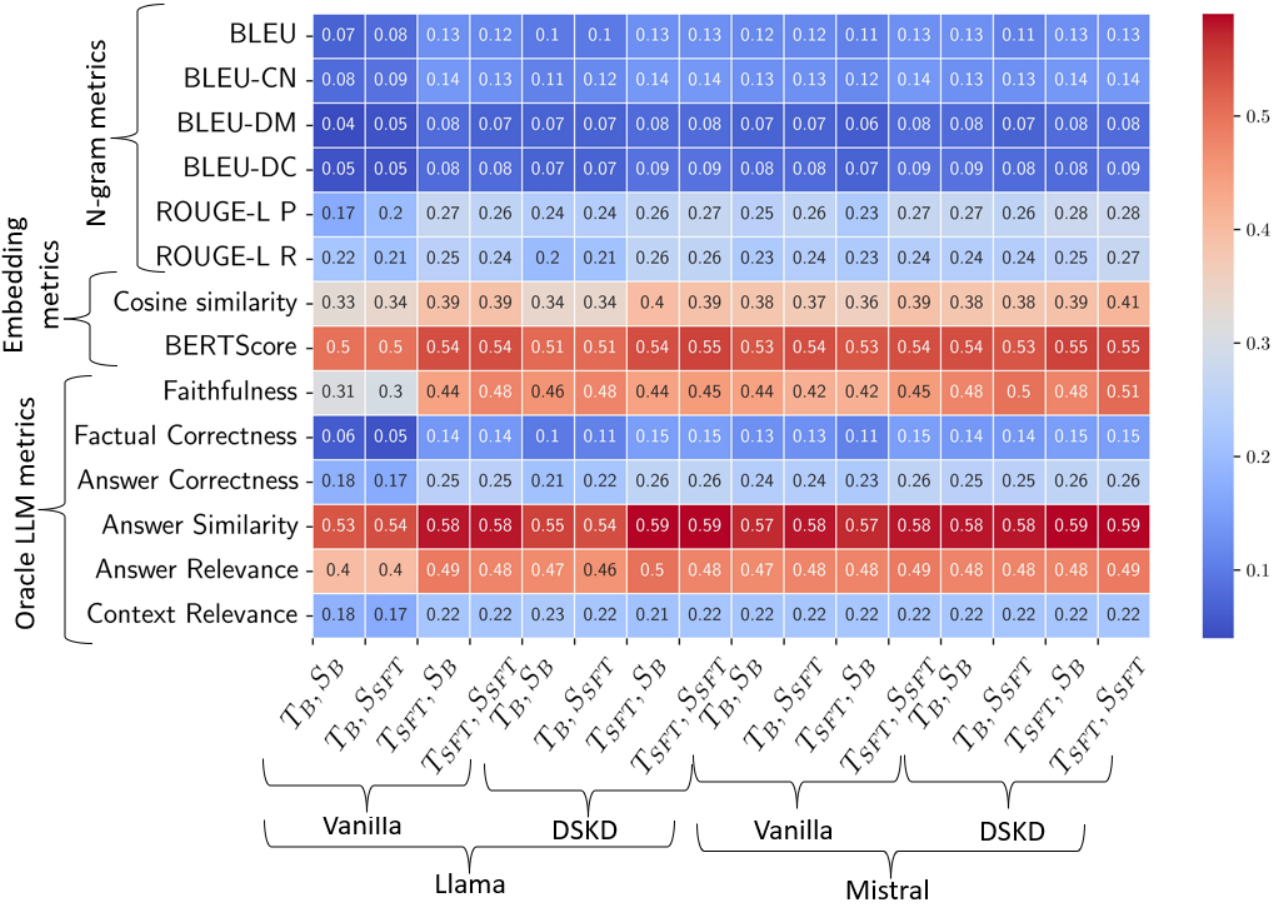}
\caption{Performance on 14 metrics for various combinations of $T_B$, $T_{SFT}$, $S_B$, $S_{SFT}$ using two KD algorithms (Vanilla and DSKD) and models of different and same vocabulary (Mistral and Llama).}
\label{fig:heatmap}
}
\end{figure*}

This is considered for both the vocabulary scenarios, ($M$, $L$), and scenario of best performing SFT models $(T_{SFT}, S_{SFT})$ and untrained models $(T_{B}, S_{B})$; the latter accounts for scenarios when training models is not feasible. This results in 4 NH tests for each of the 14 metrics. 
\end{itemize}
\vspace{-2mm}
\subsection{Dataset description}\label{subsec:dataset}
We consider samples from TeleQuAD {\cite{telequad2025}}, a Telecom QA dataset, curated using publicly available 3GPP (Rel 15) documents  \cite{3gpp_release_15}. The training, development and test data comprise of 2385, 726 and 597 QA pairs, respectively, derived from 452 contexts (sections). 

\subsection{Environment}\label{subsec:environment}
In our experiments, we have considered Llama-7b\footnote{\url{https://huggingface.co/meta-llama/Llama-2-7b}}, Mistral-7b\footnote{\url{https://huggingface.co/mistralai/Mistral-7B-v0.1}} as teacher models and Tinyllama-1.1b\footnote{\url{https://huggingface.co/TinyLlama/TinyLlama-1.1B-Chat-v1.0}} as student model. 
The GPU used for training and inference is NVIDIA A100-SXM4-80GB. Table \ref{tab:parameters} summarizes the parameters considered for $T_{SFT}$ and $S_{SFT}$.

\begin{table}[h]
    \centering
    \begin{tabular}{|p{3.6cm} | p{4cm}|}
    \hline
         Model source & Huggingface model hub\footnote{\href{https://huggingface.co/}{Hugging Face Model Repository}}\\ \hline
         Maximum epoch & 50\\ \hline
         Early stopping criteria & minimum improvement + 0.01\\ \hline
         Early stopping patience & 3 epochs \\ \hline
         Learning rate & 0.001, cosine decay\\ \hline
         SFT algorithm & Low-Rank Adaptation (LoRA) \cite{hu2021lora}\\ \hline
         Rank & 256\\ \hline
         Alpha & 8\\ \hline
         Dropout & 0.1\\ \hline
    \end{tabular}
    \caption{Summary of the parameters for SFT.}
    \label{tab:parameters}
\end{table}

\section{Experimental Results}
\label{sec:ExperimentalResults}


\begin{figure*}[h!]
\centering
\includegraphics[width=1.01\textwidth]{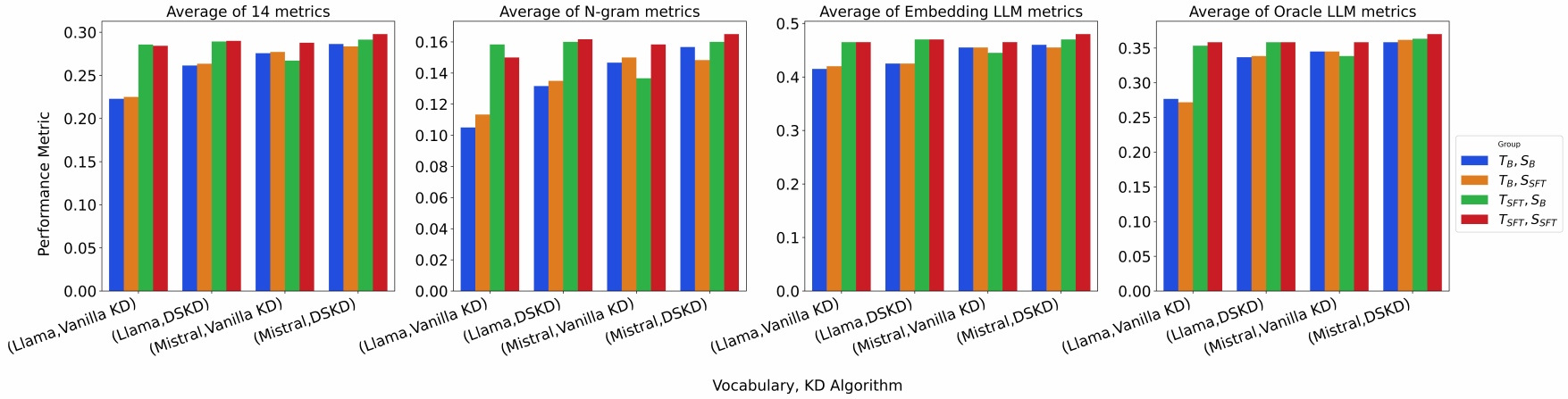}
\caption{Group-wise average of performance metrics from the heatmap in Fig.~\ref{pic:heatmap}.}
\label{fig:avg_metrics}
\end{figure*}

Fig. \ref{fig:heatmap} shows the heatmap depicting performance of 16 combinations of KD for 14 metrics. For brevity, we also report the mean of all 14 metrics and group-wise metrics (N-gram metrics, embedding based metrics and Oracle-LLM metrics) in Fig. \ref{fig:avg_metrics}.

We systematically analyze the results and organize our findings as impact of (i) SFT (RQ1) (ii) SFT on teacher and student (RQ1) (iii) vocabulary and KD algorithm (RQ2) (iv) performance metrics groups (RQ3)
\vspace{-2mm}
\subsection{Impact of SFT}
We organize analysis with vocabulary as starting point: 
\subsubsection{Llama}
Consider the  bar plots which depicts Llama as the teacher in Fig. \ref{fig:avg_metrics} i.e., the bars denoting (Llama, Vanilla KD) and (Llama, DSKD). We observe that SFT of teacher/student/both results in improvement of performance irrespective of the training algorithm (first bar vs the subsequent 3 bars). 
The improvement is statistically significant (refer to $H_{train}^{S}$, $H_{train}^{T}$, $H_{train}^{T,S}$ in Table \ref{tab:htests}). Here, we observe that NH is rejected for most metrics (13 out of 14 for Vanilla KD and 8 or 9 out of 14 for DSKD) with SFT of student or teacher or both for Llama vocabulary, irrespective of algorithms. From this and from the average performance metrics, we infer that { SFT results in statistically significant performance improvement when we choose models of same vocabulary (Llama and TinyLlama pair) models}.
\subsubsection{Mistral}
Refer to Mistral set of bar plots in Fig \ref{fig:avg_metrics} i.e., the bars denoting (Mistral, Vanilla KD) and (Mistral, DSKD). We observe that training improves results Fig. \ref{fig:avg_metrics}, but the improvement is not statistically significant (refer row Mistral in Table \ref{tab:htests}); NH is not rejected $H_{train}^{S}$, $H_{train}^{T}$, $H_{train}^{T,S}$ i.e., 0 out of 14 metrics are NH rejected. 

Thus, combining these findings, we infer that \textbf{training using SFT improves performance across vocabulary and algorithms; improvement with SFT of teacher and/or student when models have same vocabulary is significant}.
\vspace{-2mm}
\subsection{Impact of SFT on teacher and student }

\subsubsection{Llama}
We observe from Llama set of bar plots of Fig \ref{fig:avg_metrics} (i.e., the bars denoting (Llama, Vanilla KD) and (Llama, DSKD)), that SFT results in improved performance for ($T_{SFT}$, $S_B$) and ($T_{SFT}$, $S_{SFT}$) than for ($T_{B}$, $S_{SFT}$) across metric groups. From Table \ref{tab:htests}, Llama row and $H_{SFT}^S$, we see that NH is rejected for most cases irrespective of the KD algorithm - improvement of ($T_{SFT}$, $S_{SFT}$) over ($T_{B}$, $S_{SFT}$) is more significant than that of ($T_{SFT}$, $S_{SFT}$) over ($T_{SFT}$, $S_{B}$) combination i.e., NH is rejected in 13 and 9 out 14 metrics for $H_{SFT}^S$ and not rejected for any metric in $H_{SFT}^T$. This implies that \textbf{SFT of teacher model before KD is useful and it is not necessary to train both teacher and student when choosing  models of same vocabulary}. 

\subsubsection{Mistral}
When vocabulary is different i.e., refer Mistral set of results in Fig \ref{fig:avg_metrics}, the bars denoting (Mistral, Vanilla KD) and (Mistral, DSKD), we observe that best performance seen in ($T_{SFT},S_{SFT}$), followed by ($T_{B},S_{SFT}$), and an apparent dip in performance is observed for ($T_{SFT},S_{B}$) or ($T_{B},S_{SFT}$). Referring to Table \ref{tab:htests} for Mistral, all the H-SFT tests show that NH is not rejected for any of the metrics for both $H_{SFT}^T$ and $H_{SFT}^S$. This implies \textbf{the performance improvement/dip with both models trained or either model trained is not statistically significant irrespective of the KD algorithm}. We suspect this could be because of limited training data and one of the potential future work direction could be towards SFT results with more training samples. 

\begin{table}[t!]
    \centering
    \renewcommand{\arraystretch}{1.5}
\resizebox{0.5\columnwidth}{!}{%
    \begin{tabular}{ |c|cc|cc|}
    \hline
       Algorithm & Llama-V  &  Llama-D & Mistral-V  & Mistral-D \\ \hline
        $H_{train}^{S}$ & 0&1 & 0&0 \\
        $H_{train}^{T}$ & 13&9 & 0&0 \\
        $H_{train}^{T,S}$ & 13& 8& 0&0 \\  \hline
        $H_{SFT}^{S}$ & 13& 9& 0&0 \\
        $H_{SFT}^{T}$ & 0& 0& 0&0 \\\hline 
        $H_{Alg}^{T,S}$ & \multicolumn{2}{c|}{0} & \multicolumn{2}{c|}{1} \\
        $H_{Alg}^{B}$ & \multicolumn{2}{c|}{10} & \multicolumn{2}{c|}{14} \\
         \hline
    \end{tabular}
}
    \caption{Count of metrics for which NH is rejected for each of the hypotheses listed in Section \ref{ssub:hypothesis_tests}}
    \label{tab:htests}
\end{table}


\subsection{Impact of KD algorithm and Vocabulary}
\subsubsection{Algorithm} Consider the best performing pair of models i.e., ($T_{SFT}, S_{SFT}$). We observe that the model performance improvement exists, but is not significant enough; refer to Table \ref{tab:htests} where NH is not rejected for most of the metrics for $H_{Alg}^{T,S}$, across vocabularies (NH rejected is 0 out of 14 for Llama and 1 out of 14 metrics for Mistral model). \textbf{This implies, KD performance doesn't depend on the algorithm choice with SFT of both teacher and student. However, when SFT is not feasible, we observe that performance is (statistically) better with DSKD algorithm } because NH is rejected for most of the metrics for (10 out of 14 and 14 out of 14) vocabulary with $H_{Alg}^B$. 
\subsubsection{Vocabulary} Consider barplots corresponding to ($T_{SFT}, S_{SFT}$) in Fig. \ref{fig:avg_metrics}, i.e., SFT of both teacher and student models. We observe that average of all of the metrics are similar. Hence, \textbf{vocabulary does not impact performance with both models are trained.} However, \textbf{when training is not a feasibility, using models from different vocabulary with DSKD algorithm shows better performance. }
\vspace{-1mm}
\subsection{Impact of Performance Metrics}
From Fig. \ref{fig:avg_metrics} and Table  \ref{tab:htests}, we observe that although the range of results of the 3 groups of metrics are different, the trends followed across groups are in alignment. \textbf{There is no metric group where the results are contradictory. }
\subsection{Summary}
We summarize the findings from the results section above here.
\begin{itemize}
    \item RQ1
    \begin{itemize}
        \item Training teacher models, student models or both using SFT improves performance across Llama, Mistral models and Vanilla KD and DSKD algorithms.
        \item When teacher and student  models are Llama and tinyLlama, SFT of teacher before KD is useful and it may not be necessary to train both teacher and student.
        \item When teacher model is Mistral, and student model is TinyLlama, the performance improvement with SFT of both teacher and student models is not statistically significant over that of either model being trained, and this holds irrespective of the KD algorithm.
    \end{itemize}
    \item RQ2
    \begin{itemize}
        \item If one performs SFT of both teacher and student, distilled model performance doesn't depend on the algorithm or teacher model (Llama or Mistral) choice. 
        \item When SFT is not feasible, we observe that performance is (statistically) better using Mistral as teacher model, TinyLlama as student model with DSKD algorithm. 
    \end{itemize}
    \item RQ3
    \begin{itemize}
        \item The performance results follow similar trends across metric groups. 
    \end{itemize}
\end{itemize}

\section{Conclusions \& Future Work}\label{sec:conclusions}
In this work, we have systematically studied impact of SFT of teacher and student model prior to KD of LLMs from perspective of vocabulary match, KD algorithms, variants of teacher SFT or student SFT or both. Our results are based on a telecom QA dataset and we use various metrics for an overall perspective. We have discussed outcome of RQs through performance results and statistical tests. From our analysis, we recommend that when teacher is Mistral, training using SFT improves performance across vocabulary and algorithms; improvement with SFT of teacher and/or student when models have same vocabulary is significant. When Llama is the teacher, SFT of teacher model before KD is useful and it is not necessary to train both teacher and student when choosing models of same vocabulary. 

Future work would involve extending it to other tasks like code generation and agent-based systems. Another direction for future work is towards model size - the teacher models used in this work are of relatively smaller size. Evaluation of KD from larger models including Mixture of Experts (MoE) models for domain-specific tasks would be important for the community.
\bibliographystyle{IEEEtran}
\bibliography{citation}

\begin{thebibliography}{10}
\providecommand{\url}[1]{#1}
\csname url@samestyle\endcsname
\providecommand{\newblock}{\relax}
\providecommand{\bibinfo}[2]{#2}
\providecommand{\BIBentrySTDinterwordspacing}{\spaceskip=0pt\relax}
\providecommand{\BIBentryALTinterwordstretchfactor}{4}
\providecommand{\BIBentryALTinterwordspacing}{\spaceskip=\fontdimen2\font plus
\BIBentryALTinterwordstretchfactor\fontdimen3\font minus \fontdimen4\font\relax}
\providecommand{\BIBforeignlanguage}[2]{{%
\expandafter\ifx\csname l@#1\endcsname\relax
\typeout{** WARNING: IEEEtran.bst: No hyphenation pattern has been}%
\typeout{** loaded for the language `#1'. Using the pattern for}%
\typeout{** the default language instead.}%
\else
\language=\csname l@#1\endcsname
\fi
#2}}
\providecommand{\BIBdecl}{\relax}
\BIBdecl

\bibitem{soman2023observations}
S.~Soman and H.~G. Ranjani, ``Observations on {LLMs} for telecom domain: capabilities and limitations,'' in \emph{Proceedings of the Third International Conference on AI-ML Systems}, 2023, pp. 1--5.

\bibitem{bariah2023understanding}
L.~Bariah, H.~Zou, Q.~Zhao, B.~Mouhouche, F.~Bader, and M.~Debbah, ``Understanding telecom language through large language models,'' in \emph{GLOBECOM 2023-2023 IEEE Global Communications Conference}.\hskip 1em plus 0.5em minus 0.4em\relax IEEE, 2023, pp. 6542--6547.

\bibitem{roychowdhury2024evaluation}
S.~Roychowdhury, S.~Soman, H.~G. Ranjani, N.~Gunda, V.~Chhabra, and S.~K. Bala, ``Evaluation of {RAG} metrics for question answering in the telecom domain,'' in \emph{ICML 2024 Workshop on Foundation Models in the Wild}, 2024.

\bibitem{thanos2024}
A.~Karapantelakis, M.~Thakur, A.~Nikou, F.~Moradi, C.~Olrog, F.~Gaim, H.~Holm, D.~D. Nimara, and V.~Huang, ``Using large language models to understand telecom standards,'' in \emph{2024 IEEE International Conference on Machine Learning for Communication and Networking (ICMLCN)}, 2024, pp. 440--446.

\bibitem{zou2024telecomgpt}
H.~Zou, Q.~Zhao, Y.~Tian, L.~Bariah, F.~Bader, T.~Lestable, and M.~Debbah, ``Telecomgpt: A framework to build telecom-specfic large language models,'' \emph{arXiv preprint arXiv:2407.09424}, 2024.

\bibitem{piovesan2024telecom}
N.~Piovesan, A.~De~Domenico, and F.~Ayed, ``Telecom language models: Must they be large?'' \emph{arXiv preprint arXiv:2403.04666}, 2024.

\bibitem{maatouk2024large}
A.~Maatouk, N.~Piovesan, F.~Ayed, A.~De~Domenico, and M.~Debbah, ``Large language models for telecom: Forthcoming impact on the industry,'' \emph{IEEE Communications Magazine}, 2024.

\bibitem{schick2020s}
T.~Schick and H.~Sch{\"u}tze, ``It's not just size that matters: Small language models are also few-shot learners,'' \emph{arXiv preprint arXiv:2009.07118}, 2020.

\bibitem{zhang2023revisiting}
C.~Zhang, J.~Cheng, I.~Shumailov, G.~A. Constantinides, and Y.~Zhao, ``Revisiting block-based quantisation: What is important for sub-8-bit llm inference?'' \emph{arXiv preprint arXiv:2310.05079}, 2023.

\bibitem{ma2023llm}
X.~Ma, G.~Fang, and X.~Wang, ``Llm-pruner: On the structural pruning of large language models,'' \emph{Advances in neural information processing systems}, vol.~36, pp. 21\,702--21\,720, 2023.

\bibitem{gou2021knowledge}
J.~Gou, B.~Yu, S.~J. Maybank, and D.~Tao, ``Knowledge distillation: A survey,'' \emph{International Journal of Computer Vision}, vol. 129, no.~6, pp. 1789--1819, 2021.

\bibitem{xu2024survey}
X.~Xu, M.~Li, C.~Tao, T.~Shen, R.~Cheng, J.~Li, C.~Xu, D.~Tao, and T.~Zhou, ``A survey on knowledge distillation of large language models,'' \emph{arXiv preprint arXiv:2402.13116}, 2024.

\bibitem{Hinton2015DistillingTK}
\BIBentryALTinterwordspacing
G.~E. Hinton, O.~Vinyals, and J.~Dean, ``Distilling the knowledge in a neural network,'' \emph{ArXiv}, vol. abs/1503.02531, 2015. [Online]. Available: \url{https://api.semanticscholar.org/CorpusID:7200347}
\BIBentrySTDinterwordspacing

\bibitem{vaswani2017attention}
A.~Vaswani, N.~Shazeer, N.~Parmar, J.~Uszkoreit, L.~Jones, A.~N. Gomez, {\L}.~Kaiser, and I.~Polosukhin, ``Attention is all you need,'' \emph{Advances in neural information processing systems}, vol.~30, 2017.

\bibitem{kolesnikova2022knowledge}
A.~Kolesnikova, Y.~Kuratov, V.~Konovalov, and M.~Burtsev, ``Knowledge distillation of russian language models with reduction of vocabulary,'' \emph{arXiv preprint arXiv:2205.02340}, 2022.

\bibitem{aguilar2020knowledge}
G.~Aguilar, Y.~Ling, Y.~Zhang, B.~Yao, X.~Fan, and C.~Guo, ``Knowledge distillation from internal representations,'' in \emph{Proceedings of the AAAI conference on artificial intelligence}, vol.~34, no.~05, 2020, pp. 7350--7357.

\bibitem{zhang2024dual}
S.~Zhang, X.~Zhang, Z.~Sun, Y.~Chen, and J.~Xu, ``Dual-space knowledge distillation for large language models,'' in \emph{Proceedings of the 2024 Conference on Empirical Methods in Natural Language Processing (EMNLP),Miami Florida USA}, 2024.

\bibitem{desmond2024evalullm}
M.~Desmond, Z.~Ashktorab, Q.~Pan, C.~Dugan, and J.~M. Johnson, ``Evalullm: Llm assisted evaluation of generative outputs,'' in \emph{Companion Proceedings of the 29th International Conference on Intelligent User Interfaces}, 2024, pp. 30--32.

\bibitem{shi2022evaluation}
E.~Shi, Y.~Wang, L.~Du, J.~Chen, S.~Han, H.~Zhang, D.~Zhang, and H.~Sun, ``On the evaluation of neural code summarization,'' in \emph{Proceedings of the 44th international conference on software engineering}, 2022, pp. 1597--1608.

\bibitem{lin-2004-rouge}
\BIBentryALTinterwordspacing
C.-Y. Lin, ``{ROUGE}: A package for automatic evaluation of summaries,'' in \emph{Text Summarization Branches Out}.\hskip 1em plus 0.5em minus 0.4em\relax Barcelona, Spain: Association for Computational Linguistics, Jul. 2004, pp. 74--81. [Online]. Available: \url{https://aclanthology.org/W04-1013}
\BIBentrySTDinterwordspacing

\bibitem{reimers2020making}
N.~Reimers and I.~Gurevych, ``Making monolingual sentence embeddings multilingual using knowledge distillation,'' \emph{arXiv preprint arXiv:2004.09813}, 2020.

\bibitem{zhang2019bertscore}
T.~Zhang, V.~Kishore, F.~Wu, K.~Q. Weinberger, and Y.~Artzi, ``Bertscore: Evaluating text generation with bert,'' \emph{arXiv preprint arXiv:1904.09675}, 2019.

\bibitem{es2024ragas}
S.~Es, J.~James, L.~E. Anke, and S.~Schockaert, ``Ragas: Automated evaluation of retrieval augmented generation,'' in \emph{Proceedings of the 18th Conference of the European Chapter of the Association for Computational Linguistics: System Demonstrations}, 2024, pp. 150--158.

\bibitem{telequad2025}
\BIBentryALTinterwordspacing
F.~Gebre, H.~Holm, M.~Gunnarsson, D.~Nimara, J.~Wei, V.~Huang, A.~Sharma, and H.~G. Ranjani, ``{TeleQuAD}: A suite of question answering datasets for the telecom domain,'' 2025. [Online]. Available: \url{https://github.com/EricssonResearch/TeleQuAD}
\BIBentrySTDinterwordspacing

\bibitem{gehan1965generalized}
E.~A. Gehan, ``A generalized wilcoxon test for comparing arbitrarily singly-censored samples,'' \emph{Biometrika}, vol.~52, no. 1-2, pp. 203--224, 1965.

\bibitem{3gpp_release_15}
3GPP, ``{3GPP} release 15,'' \url{https://www.3gpp.org/specifications-technologies/releases/release-15}, 3GPP, Tech. Rep., 2019, accessed: 2024-05-19.

\bibitem{hu2021lora}
E.~J. Hu, Y.~Shen, P.~Wallis, Z.~Allen-Zhu, Y.~Li, S.~Wang, L.~Wang, and W.~Chen, ``Lora: Low-rank adaptation of large language models,'' \emph{arXiv preprint arXiv:2106.09685}, 2021.

\end{thebibliography}
\end{document}